\documentclass[10pt,twocolumn,letterpaper]{article}

\usepackage{iccv}
\usepackage{times}
\usepackage{epsfig}
\usepackage{graphicx}
\usepackage{amsmath}
\usepackage{amssymb}

\usepackage{booktabs}
\usepackage{multirow}
\usepackage[dvipsnames]{xcolor}
\usepackage{soul}
\usepackage[normalem]{ulem}
\usepackage[pagebackref=true,breaklinks=true,letterpaper=true,colorlinks,bookmarks=false]{hyperref}
\usepackage{pifont}

\usepackage[breaklinks=true,bookmarks=false]{hyperref}
\iccvfinalcopy 

\def\iccvPaperID{****} 

\ificcvfinal\fi

\begin{document}
	
	\title{Exploiting Scene Graphs for Human-Object Interaction Detection}
	\author{Tao He$^1$, Lianli Gao$^2$, Jingkuan Song$^2$, Yuan-Fang Li$^1$ \thanks{Corresponding author.} \\ 
		$^1$Faculty of Information Technology, Monash University \\
		$^2$Center for Future Media, University of Electronic Science and Technology of China\\
		{\tt\small \{tao.he,yufang.li\}@monash.edu, lianli.gao@uestc.edu.cn,jingkuan.song@gmail.com} 
	}
	
	\maketitle
	\ificcvfinal\thispagestyle{empty}\fi
	
	\begin{abstract}
		Human-Object Interaction (HOI) detection is a fundamental visual task aiming at localizing and recognizing  interactions between humans and objects. 
		Existing works focus on the visual and linguistic features of the humans and objects. {However, they do not capitalise on the high-level and semantic {relationships} present in the image, which provides crucial contextual and {detailed relational knowledge} for HOI inference.} 
		We propose a novel method to exploit this information, through the scene graph, for the Human-Object Interaction (SG2HOI) detection task. 
		Our method, SG2HOI, incorporates the SG information in two ways: (1) we embed a scene graph into a global context clue, serving as the scene-specific environmental context; and (2) we build a relation-aware message-passing module to gather relationships from objects' neighborhood  and transfer them into interactions. Empirical evaluation shows that our SG2HOI method outperforms the state-of-the-art methods on two benchmark HOI datasets: V-COCO and HICO-DET. Code will be available at \url{https://github.com/ht014/
			SG2HOI}.
	\end{abstract}
	
\section{Introduction}
Recently, Human-Object Interaction (HOI) detection \cite{gkioxari2018detecting,liao2020ppdm} aims at detecting the types of interactions of human-object pairs. It has gained increasing attention in the computer vision community as it has a wide range of practical applications, e.g.,\ action recognition~\cite{jhuang2013towards} and Human-Computer Interaction (CHI) \cite{harper2008human}. 
Formally, the goal of HOI is to detect and localize all the interaction triples in an image, i.e.\ $<$human, interaction, object$>$. HOI is a challenging problem---an image typically contains multiple humans and objects in a complex scene, while the majority of all the human-object pairs are non-relation. Therefore, some works~\cite{ulutan2020vsgnet,wang2020learning,wang2020contextual} that are solely based on visual features cannot learn good discriminative patterns for HOIs.  

It is intuitive to turn to external knowledge, e.g.,\ the well-known knowledge graph ConceptNet~\cite{speer2017conceptnet}, to mitigate the limitation of visual appearance features. However, as such knowledge graphs are often general-purpose, most of retrieved results are redundant. Thus, such external knowledge may not provide sufficiently informative cues for HOIs. Instead, we turn to another closely-related task, i.e.\ Scene Graph Generation (SGG), to generate a tiny relation (knowledge) graph for each image, serving as the external knowledge to make up the visual cues. 

\begin{figure}[]
	\centering
	\includegraphics[width=0.99\linewidth]{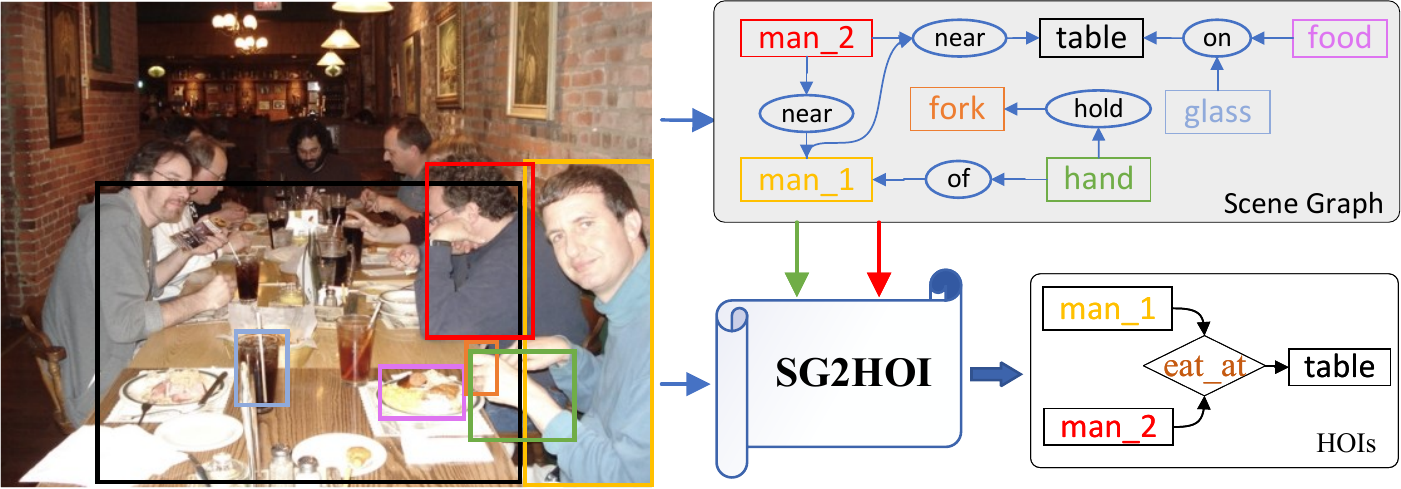}
	\caption{An illustration of our Scene Graph to Human-Object Interaction method, where we consider  the scene graph between objects as external knowledge to facilitate the prediction of HOIs in two ways: scene graph embedding (green arrow) described in Sec.~\ref{sec:sge}    and relation-aware message passing (red arrow) described in Sec.~\ref{sec:ramp}. }
	\label{fig.1}
\end{figure}



SGG~\cite{lu2016visual,pgae,xu2017scene} and HOI~\cite{gkioxari2018detecting,wang2020learning} both aim at identifying spatial and other types of relations between objects in an image. 
Figure~\ref{fig.1} illustrates a scene graph and the HOI graph of the same image. 
There are two main differences between the two tasks: (1) in SGG, subjects can be of any type (humans, cars, etc.), while in HOI they are fixed as humans, which results in more edges in the SG; and (2) the predicates of HOI only consist of interaction verbs, while in SGG many types of relations may exist, including locative prepositions (e.g.,\ \texttt{on}) and {semantic actions (e.g.,\ \texttt{play with})}. 
Simply speaking, an SG is a general relation graph whereas an HOI graph is a human-focused subject graph, which can be considered a subgraph extracted from the SG. Thus, we believe that the scene graph can provide more detailed cues for HOI detection and improve the performance of an HOI model. 
Specifically, incorporating the SG in the HOI task has two benefits: (1) the SG puts each object and human in a relation graph, which can provide contextual cues and benefit the scene understanding; and (2) information contained in relationships in the SG can be explicitly or implicitly transferred to more accurately identify the correct interactions. 
Therefore, our motivation is to develop a decoding method that transfers the knowledge encoded in the scene graph to the HOI graph. To this end, we decode an SG from two aspects: the global scene-level and regional relation-level.

For the scene-level, our goal is to extract the scene-specific cues from the detailed scene graph, because many previous works \cite{liu2020amplifying,ulutan2020vsgnet,wang2020contextual} have demonstrated that the interactions are scene-biased, that is, some interactions are highly correlated to specific scenes. 
Taking Figure~\ref{fig.1} as an example, the image is a restaurant scene from the visual appearance, and it is more likely that the interaction is about \texttt{eating}. Therefore, accurate scene cues can benefit HOI recognition. Many existing works~\cite{liu2020amplifying,ulutan2020vsgnet} have employed the global visual appearance as the scene clues. However, due to the coarseness of the appearance feature, their performance does not gain a significant improvement. Instead, we treat a scene graph as the scene-specific contextual cue and propose two components to embed it: scene graph layout encoding and attention-based relation fusion. 

%

For the relation-level, we observe that the relations in the scene graph can explicitly or implicitly transfer to interactions. For instance, as shown in Figure~\ref{fig.1}, since the three relation triples: $<$hand, of, man\_1$>$, $<$hand, hold, fork$>$, and $<$man\_1, near, table$>$ simultaneously take place, we are more likely to infer that man\_1's interaction is \texttt{eat\_at}. Besides, based on the relation $<$man\_2, near, man\_1 $>$, we could hypothesise that both of them probably have the similar interaction. Furthermore, based on his visual features, we could infer that the man is also eating at the table. Thus, the knowledge of the exact relations between object pairs makes the inference more certain.
To this end, we develop a relation-aware message passing module to reason on the SG by gathering relation information from inter- and intra- class neighbors and refine their features.  

In summary, our contribution is three-fold:
\begin{itemize}
	\item We propose a novel Scene Graph to Human-Object Interaction (\textbf{SG2HOI}) detection network to bridge the gap between the two tasks. 
	To the best of our knowledge, we are the first to utilize  scene graphs for HOI detection.
	
	\item We design two components to decode the SG: scene graph embedding and   relation-aware message-passing, to learn the environmental context  and  transfer SG relations to HOI interactions, respectively. 
	
	\item  We evaluate our approach  on two popular HOI detection benchmark datasets: V-COCO and HICO-DET, in terms of a wide range of evaluation metrics. Our evaluation shows that SG2HOI method outperforms state-of-the-art models on both two datasets. 
\end{itemize}

\section{Related Work}\label{sec:related}

\textbf{Scene Graph Generation (SGG)} \cite{xu2020survey} is to detect visual relationship of all the objects pairs in an image. This task has been widely studied for many years \cite{dai2017detecting,lu2016visual,newell2017pixels,xu2017scene,zhang2017visual}. Different from HOI,   SGG tries to detect relation triples, i.e. $<$subject, predicate, object$>$, where the subject is not limited to human, and therefore the combinations of the relation triples are much more versatile than HOI and the long-tail becomes the challenging problem in SGG. A couple of recent works~\cite{he2020learning,tang2020unbiased,zellers2018neural} proposed   a series of techniques to  address the  imbalanced distribution problem.
\cite{zellers2018neural} first observed that the biased relation distribution and    statistically showed the frequencies of predicates  are long-tailed, and even using the frequency information as the prior can obtain stunning performance. Subsequently,  \cite{tang2020unbiased} analyzed the cause effect in the inference stage and proposed a novel debiased strategy able to be seamlessly applied in other SGG models.    \cite{he2020learning} tackled this problem by feature hallucination  and  knowledge  transferring from the head  to the tail.

\begin{figure*}
	\centering
	\includegraphics[width=0.92\linewidth]{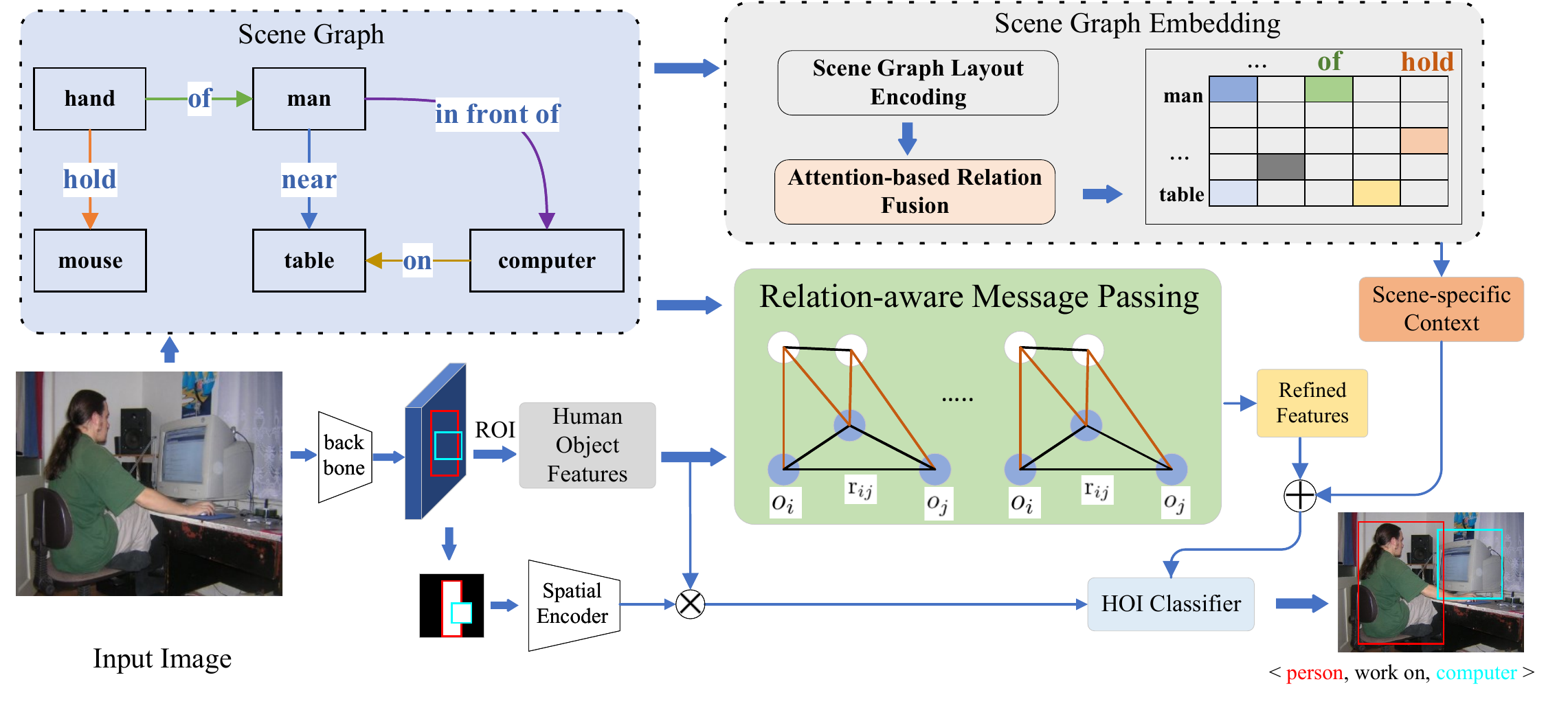}
	\caption{The overview of our Scene Graph to Human-Object Interaction (SG2HOI) method, where the scene graph is fed into two important modules: scene graph embedding and relation-aware message passing. The former aims to learn the scene-specific contextual cues while the latter aims to reason on the scene graph and gather interaction information from { neighbors}. }
	\label{framework}
\end{figure*}

\textbf{Human-Object Interaction (HOI) Detection} aims at detecting and localizing the interactions of human-object pairs, and requires a deeper and detailed understanding of the scene. Generally, most of the previous works \cite{gao2018ican,gkioxari2018detecting,gupta2019no,liao2020ppdm,qi2018learning,ulutan2020vsgnet} consistently consists of  two steps. The first is to utilize a pre-trained object detection network, e.g., Faster-RCNN~\cite{ren2016faster}, to generate all the human and object proposals and construct quadratic number of human-object pairs, and the second step feeds those pairs to  an interaction classification module. Due to the fact that the first stage is based on the off-the-shelf model, numerous works focus on the second stage and dedicate to exploring more visual and contextual information  so that the interaction classifier can capture the essential hidden relationship patterns between humans and objects. Specifically, InterPoint \cite{wang2020learning} proposed a fully convolutional approach to simultaneously detect the interaction points and predict the interactions, which avoided the computation of all the human-object pairs and improve the computational efficiency. VSGNet \cite{ulutan2020vsgnet} concentrated on the relative spatial and structural cues by deploying two modules: spatial attention network and interactiveness graph. 
TIK~\cite{li2019transferable} proposed an interactiveness network to suppress the non-interactive human-object pairs and improved HOI models' performance.  PD-Net~\cite{zhong2020polysemy} devised a Polysemy Deciphering Network to address the diverse semantic meanings of verbs. CHGNet~\cite{wang2020contextual} developed a homogeneous graph network to conduct message passing between homogeneous entities and heterogeneous entities, but their passing messages are transferred in an agnostic way.  
To address  the insufficient  and indistinguishable   visual appearance feature  for different interactions,  FCNNet \cite{liu2020amplifying} proposed a multi-stream pipeline to amplify key cues, such as object labels' word2vec, fine-grained spatial layout and  flow prediction.   

However, although SGG and HOI are different from the concentration, that is, SGG focuses more on relationship of different object pairs but HOI aims at the high-level relation (interaction),   no currently existing works   link them together. To some extent, we believe that the SG can be an important cue for HOI by supplying scene contextual information and reasoning knowledge graph.

\section{Method}

In this section, we present our Scene Graph to Human-Object Interaction (SG2HOI) framework, as shown in Figure \ref{framework}.  
Our SG2HOI framework leverages the knowledge contained in the SG for HOI detection with two main modules: \emph{scene graph embedding} (Sec.~\ref{sec:sge}) and \emph{relation-aware message passing} (Sec.~\ref{sec:ramp}), which aim at learning scene-specific contextual cues and transferring SG relationships to human-object interactions, respectively. 
Similar to prior works~\cite{ulutan2020vsgnet,wang2020contextual}, we also extract a human-object spatial map as the auxiliary feature. Finally, we fuse these features: visual appearance features of humans and object (extracted with Faster-RCNN \cite{ren2016faster}), scene graph embeddings, refined human and object features by message passing on the scene graph, as well as the spatial map to predict HOIs (Sec.~\ref{sec:hoip}).



\subsection{Scene Graph Embedding}\label{sec:sge}
Learning contextual knowledge from a scene graph can provide the model with scene-specific cues, which play an important role in HOI detection. Different from visual appearance features used in prior art~\cite{liu2020amplifying,ulutan2020vsgnet}, scene graphs can provide more high-level relationship information. For instance, in Figure \ref{fig.1},  even with the input image removed, i.e. without the visual appearance feature, and only given the SG, we could infer the interaction is highly related to \texttt{eating}, according to the relations $<$food, on, table$>$ and $<$hand, hold, fork$>$, which indicate a dining environment. To embed scene graphs, we design two components: scene graph layout encoding and relation fusion. 

\subsubsection{Scene Graph Layout Encoding}

\begin{figure}[htb]
	\centering
	\includegraphics[width=0.8\linewidth]{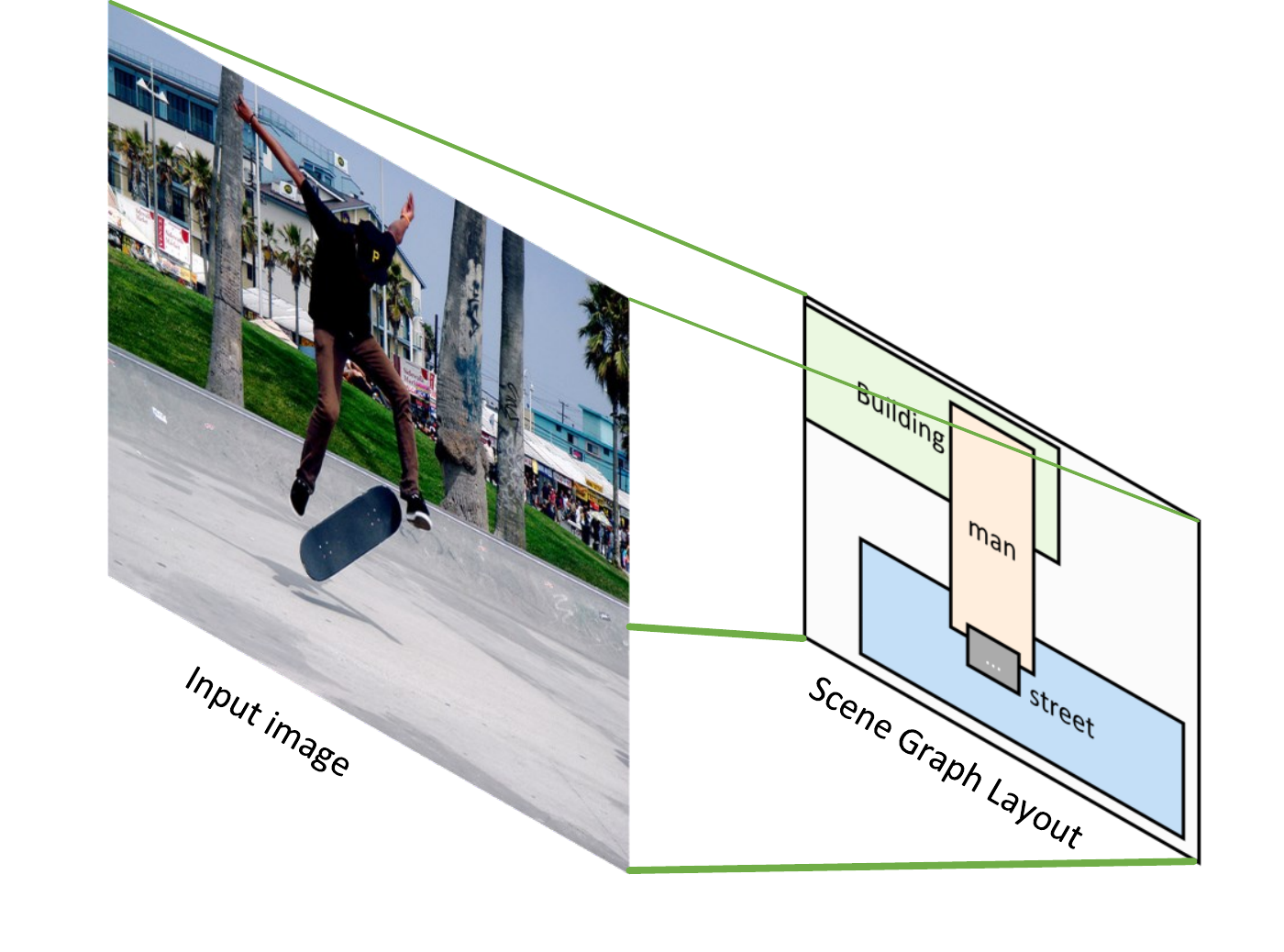}
	\caption{The scene graph layout mainly consists two items:   spatial localization and size,  and the semantic context of objects. 
	}
	\label{fig.2}
\end{figure}

The scene graph layout, as illustrated in Figure~{\ref{fig.2}}, consists of a set of objects with associated information, including their labels, sizes, and positions, as well as their relative spatial relationship. 
A scene graph layout contains two types of important cues: the \emph{spatial cues} from each object's position and size, and the \emph{contextual cues} from the objects' environment, which can be interpreted as the co-occurrence of the objects in a specific scene. For example, table, food and forks usually occur in the dining environment.


However, as HOI datasets have no scene graph annotations, we first use a model~\cite{tang2020unbiased} pre-trained on Visual Genome~\cite{krishna2017visual} to generate the scene graph $\mathcal{S}$ of each image, where  $\mathcal{S}\!\!=\!(V,E), V\!\!\!=\!\!\{o_i\}_{i=1}^n, E\!\!=\!\{e_k\!\!= <\!o_i,r_{ij},o_j\!>\!\}_{k=1}^{m}$, where $o_i$, $o_j$ are the detected objects and $r_{ij}$ is the relationship between $o_i$ and $o_j$, $n$ is the number of detected objects, and $m$ is the number of detected relations. 

After generating $\mathcal{S}$, we extract the spatial location of each object. 
Specifically, given the bounding box $b_i=\{x_{i}^t, y_i^t, x_i^b, y_i^b\}$ of the object $o_i$ where the former two dimensions represent  the top left  coordinate while the last two are the bottom right,  we define the spatial feature of $o_i$ as:
\begin{equation}
\mathrm{\mathbf{p}}_{i}  = \mathbf{W}^s   \left[ x_{i}^t, y_i^t, x_i^b, y_i^b, x_i^c, y_i^c, x_i^w, y_i^h \right]
\end{equation} 
where $(x_i^c, y_i^c)$    are the center coordinate, and $(x_i^w, y_i^h)$ are the width and height of the bounding box $b_i$; and  $\mathbf{W}^s$  denotes a transformation layer to project the original 8-D vector to   a high-dimensional representation, following~\cite{ulutan2020vsgnet}.   

Then, we extract the language prior, the \textsc{GloVe} word embeddings~\cite{pennington2014glove}, as the semantic feature {$\textbf{u}_i$}  for object $o_i$. Thus, for a object (node) in the scene graph, we concatenate the two features, denoted as $\mathbf{v}_i=[\mathrm{\mathbf{p}}_i; \mathrm{\mathbf{u}}_i]$. 

To encode the context of objects, we view the scene graph nodes as a sequence of semantic and spatial codewords: $[\mathbf{v}_1,\mathbf{v}_2, \ldots, \mathbf{v}_n]$  and employ an RNN to encode the hidden representation for each word as: 

\begin{equation}
[\mathrm{\mathbf{h}}_i]_{i=1}^n = \mathrm{RNN}([\mathbf{W}^c\mathbf{v}_i]_{i=1}^{n})
\label{eq:h}
\end{equation}
where $\mathbf{W}^c$ is a   transformation matrix and $\mathrm{h}_i$ is the node feature equipped with  contextual cues. Note that the  nodes' order in the sequence is base on by the  left-to-right central x-coordinate of each object, following \cite{zellers2018neural}  .

\begin{figure}[htb]
	\centering
	\includegraphics[width=0.9\linewidth]{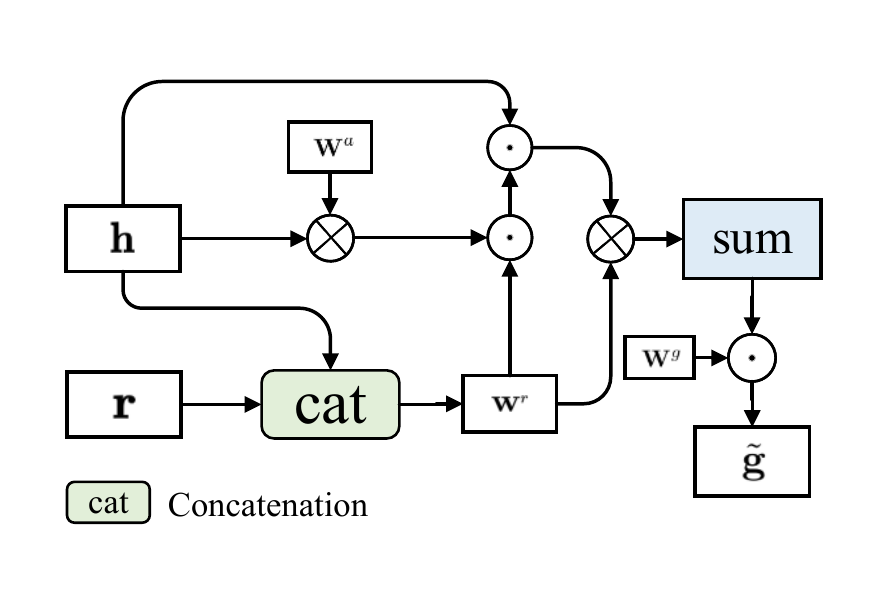}
	\caption{  Architecture of our attention-based relation fusion component, where $\mathbf{h}$ denotes   objects' contextual features derived from the scene graph layout encoding and $\mathbf{r}$ is the word embedding of relationships in the scene graph. $\odot$ and $\otimes$ are the Hadamard product and element-wise product, respectively. }
	\label{fig.3}
\end{figure}

\subsubsection{Attention-based Relation Fusion}
The relationships in the scene graph is the core component which plays an import role in the scene understanding. In this module, we dedicate to fusing the scene graph's layout and relationships into a comprehensive representation. For a given relationship $<\!\!o_i, r_{ij},o_{j}\!\!>$, we represent it as a concatenation of three features: 
\begin{equation}
\mathbf{e}_k= \mathbf{W}^r[\mathrm{\mathbf{h}}_i; \alpha_{ij}; \mathrm{\mathbf{h}}_j]\label{eqn:ek}
\end{equation}
where $\alpha_{ij}$ is the word embedding of the relation $\mathrm{{r}}_{ij}$, produced in the same way as the semantic features $\mathrm{\mathbf{v}}_i$, and $\mathbf{W}^r$ is a transform layer.

Now, for a scene graph $\mathcal{S}_i$ of the image $i$, we have its object features matrix $\mathbf{h}_i \in \mathbb{R}^{n\times d}$   and the relation features matrix $\mathbf{e}_i \in \mathbb{R}^ {m\times d}$, where $d$ is the dimension of the features. 

Before fusing $\mathbf{e}_i$ and $\mathbf{h}_i$, we first calculate a correlation matrix between them  denoting the relevance of objects on  relationships, because we believe that each object have different correlation with each relation. For instance, in Figure~\ref{fig.1}, the relation  $<$hand, hold, fork$>$  is more related to \texttt{man\_1} and \texttt{fork} than \texttt{man\_2}. 
To this end, we use a self-attention mechanism~\cite{vaswani2017attention} to calculate the correlation matrix $\mathcal{C}_i \! \in \!\mathbb{R}^{n \times m}$ as:
\begin{equation}
\mathcal{C}_i =   (\mathbf{h}_i \otimes \mathbf{W}^a) \cdot \mathbf{e}_i^\top 
\end{equation}
where $\otimes$ is the pointwise product, ${\mathbf{W}^a} $ is the learnable parameters denoting the self-attention weights, and $\cdot$ is matrix multiplication.

Then, we use $\mathcal{C}_i$ as the bridge to fuse objects and relationships together by:
\begin{equation}
\mathring{\mathbf{g}}_i =   (  {\mathcal{C}_i^{\top}} \cdot \mathbf{h}_i)  \otimes \mathbf{e}_i
\end{equation}
where $\mathring{\mathbf{g}}_i$ is the fusion of the layout and relationships {of the scene graph}. Figure~\ref{fig.2}   shows the full architecture of data flow. However, it is worth noting that the dimension of  $\mathring{\mathbf{g}}_i$ is $\mathbb{R}^{m \times d}$, and here, we use summation operation on the first dimension of $\mathring{\mathbf{g}}$ to reduce it into a vector and add a transformation layer $\mathbf{W}^g$ to embed $\mathring{\mathbf{g}}_i$  into the final representation:  
\begin{equation}
\tilde{\mathbf{g}}_i = \mathbf{W}^g \sum_{j=1}^m \mathbf{\mathring{g}}_{ij}
\end{equation}

\subsection{Relation-aware Message Passing }\label{sec:ramp}

A human-object interaction could involve several surrounding objects and thus be related to multiple relationships in the scene graph.  Thus, it is intuitive to develop a reasoning module where we can aggregate relation information from neighbors and then learn the intrinsic correlation between relationships and interactions. Toward this end, we devise a relation-aware message passing strategy able to aggregate and propagate relation signals on the scene graph and refine the visual feature of  humans and objects. 

Although previous works~\cite{qi2018learning,wang2020contextual} have proposed message passing mechanisms, the messages in these methods are {homogeneous}, i.e., they are not aware of the specific relations present in an image. This causes the gathered information to be ambiguous and unrepresentative. To address this issue, we propose to use the explicit relations in the scene graph as messages to refine the human and object visual features. We define two types of messages for feature refinement: inter- and intra-class messages, to capture the different roles humans and objects play in the interaction.


\noindent \textbf{Inter-class refinement.} For the inter-class refinement, human nodes only aggregate the messages from object (i.e.\ non-human) nodes, aimed at updating humans' feature  by  receiving relation messages from object neighbors. For example, in Figure \ref{fig.1}, by incorporating  the relation message \texttt{of} from \texttt{hand}, the model could reason that \texttt{man\_1} is holding a \texttt{fork}. 

Given the visual feature of human $\mathbf{f}_h^i$ and object  $\mathbf{f}_o^j$, we refine the human feature by aggregating the relationship messages from  its object neighbors. The inter-messages are calculated by:    
\begin{equation}
\mathring{\mathbf{f}}_h^{i}=\mathbf{W}_{o \rightarrow h}\left(\sum_{j \in \mathcal{N}^{i}}  \mathbf{W}^o \cdot \mathbf{f}_{o}^j \otimes \alpha_{ji}^{o \rightarrow h} \right)
\label{eq.7}
\end{equation}
where $\mathcal{N}^{i}$ denotes the neighbors of human $i$ in the scene graph $\mathcal{S}$; $\alpha_{ji}^{o \rightarrow h}$ is the word embedding of relation $\mathrm{r}_{ji}$ from object to human as described in Eq.~(\ref{eqn:ek}); and $\mathbf{W}^o$ and $\mathbf{W}_{o \rightarrow h}$ are two {transformation layers}. 

\noindent\textbf{Intra-class refinement.} On the other hand, a human could interact with several other humans, and their relationships could provide essential clues. Different from the inter-class refinement strategy, the intra-class messages are calculated from human neighbors only:
\begin{equation}
\overline{\mathbf{f}}_h^{i} = \mathbf{W}_{h \rightarrow h}\left(\sum_{j \in \mathcal{N}^{i}}  \mathbf{W}^h \cdot \mathbf{f}_{h}^j \otimes \alpha_{ji}^{h \rightarrow h} \right)
\label{eq.8}
\end{equation}
Then, the refined human feature is formulated as:
\begin{equation}
\tilde{ \mathbf{f}}_h^i =  \mathbf{f}_h^{i} +\overline{\mathbf{f}}_h^{i}+ \mathring{\mathbf{f}}_h^{i}
\label{eq.9}
\end{equation}

For  iterative refinement, we only let $\mathbf{f}_h^{i}$ be updated as $\tilde{ \mathbf{f}}_h^i$ and repeat the calculation following Eq.\ (\ref{eq.7}), (\ref{eq.8}), and (\ref{eq.9}).  

Note that the object feature refinement also has inter- and intra- parts and is similar to the human refinement, and we omit their details for brevity reasons. 

\subsection{HOI Predication}\label{sec:hoip}
After embedding the scene graph and refining visual features, we have the global scene graph representation $\tilde{\mathbf{g}}_k$ for the image $k$ and the refined human feature $\tilde{ \mathbf{f}}_h^i$ and object feature $\tilde{ \mathbf{f}}_o^i$. In addition, we also extract the spatial mask of each human-object pair with size $64 \times 64$ as the auxiliary feature $\mathbf{f}_s^{ij}$, following ~\cite{ulutan2020vsgnet,wang2020contextual}. 
The final prediction of HOI is combined from two branches: visual appearance features and refined features by message passing.
For the human $i$ and object $j$,  the visual branch is predicted by:

\begin{equation}
\mathbf{ {p}}_v = \mathrm{\delta}(\mathbf{W}_v[\mathbf{f}_s^{ij} \otimes [\mathbf{f}_h^{i};\mathbf{f}_o^{j}]])
\end{equation}
where $\mathrm{\delta}$ is the sigmoid active function and $\mathrm{\textbf{W}}_v$ is the classifier parameters for visual features. 

The message passing branch is calculated as below:
\begin{equation}
\mathbf{p}_m = {\delta}(\mathbf{W}_m[\tilde{\mathbf{g}}_k;\tilde{ \mathbf{f}}_h^{i};\tilde{ \mathbf{f}}_o^{j}])  
\end{equation}
where $\mathrm{\textbf{W}}_m$ is the classifier parameters for refined features.
Finally, we combine the two predicted scores  as:
\begin{equation}
\mathbf{p}^{ij} = \lambda^{ij} \cdot  \mathbf{p}_v \cdot \mathbf{p}_m
\end{equation}
where $\lambda^{ij}$ is the multiplication of the detected human and object score normalized by \cite{li2019transferable}.

In the training stage, we optimize the Binary Cross-Entropy (BCE) loss  on $\mathbf{p}^{ij}$ to optimize the HOI model.

\section{Experiments}
In this section, we first briefly describe the datasets, evaluation metrics and baseline models, as well as the implementation details. Next, we compare our model with the state-of-art methods and further conduct a series of ablation studies. Last, we show several qualitative results on the both tasks: SGG and HOI. 

\subsection{Datasets and Metrics}

\noindent\textbf{Datasets:} We evaluate our model's performance on two standard benchmarks: the V-COCO~\cite{gupta2015visual} and HICO-DET~\cite{chao2018learning} datasets. The training, validation and test sets of V-COCO consist of $2,533$, $2,867$ and $4,946$ images, respectively. Following previous works~\cite{ulutan2020vsgnet,wang2020learning}, we  also use the training and validation set, containing $5,400$ images in total, to train our model. Each human-object pair in V-COCO is labeled with a $29$-D one-hot vector. It is worth noting that among the $29$ actions, three (cut, hit, eat) have no interaction objects. HICO-Det is a large dataset and is split into two sets: $38,118$ training and $9,658$ testing images. In total, HICO-Det is annotated with $600$ human-object interaction classes, $80$ object classes and $117$ actions, including a no-interaction class.  Following previous works~\cite{chao2018learning,wang2020learning}, we categorize the interactions into three groups: Full, Rare and Non-Rare, based on the number of its training samples. Also, we further conduct the experiments under a ``Known Objects'' setting but the default  is unknown for objects. 

\noindent \textbf{Metric:}
We conduct model evaluation with the standard evaluation metric \cite{ulutan2020vsgnet,wang2020learning} role mean Average Precision ({mAP}$_\mathrm{role}$). More concretely, if an HOI triple meets the following two conditions: (1) both of bounding boxes of the detected human and object are greater than $0.5$ with the respective annotated ground-truth boxes; and (2) their interaction class is correctly predicted, then we consider the HOI triple is correct.

\noindent \textbf{Baseline Models:}
We compare our model against the following state-of-the-art models: InteractNet~\cite{gkioxari2018detecting}, GPNN~\cite{qi2018learning},  iCAN~\cite{gao2018ican}, TIK~\cite{li2019transferable}, VSGNet~\cite{ulutan2020vsgnet}, InterPoint~\cite{wang2020learning}, FCNNet~\cite{liu2020amplifying}, PD-Net~\cite{zhong2020polysemy}, and CHGNet~\cite{wang2020contextual}. A detailed discussion of these methods can be found in Sec.~\ref{sec:related}.

\subsection {Implementation Details}
For the backbone network, we use the Resnet-50 \cite{he2016deep} as the default feature extraction network for fair comparison with other models. During the training stage, we freeze the parameters of the backbone derived from the pre-trained model on ImageNet~\cite{deng2009imagenet}. The input image size is rescaled to $600 \times 800$ or $800 \times 600$, depending on the width and height of the image. We use the ROIAlign module to extract the human and object proposal features, which are transformed by two fully connected non-linear layers. In addition, the human and object bounding boxes are generated by Faster-RCNN trained on the COCO~\cite{lin2014microsoft} dataset, and the threshold score for the detected human and object are empirically set as $0.6$ and $0.3$, following \cite{ulutan2020vsgnet}. The semantic word embeddings are from \textsc{GloVe}~\cite{pennington2014glove} and all the word embedding  dimension is set to $300$.  

For the spatial feature extraction, we use two $64\times64$ masks as the input, but different from previous works~\cite{ulutan2020vsgnet,wang2020learning} that used two binary masks, we use two semantic masks, i.e., the object proposal region is filled with the corresponding object category information to differentiate them with each other. 

The scene graph is generated by the recent model~\cite{tang2020unbiased}. We train it on the Visual Genome dataset~\cite{krishna2017visual}, which is a large dataset for visual relationship detection, with more than $100,000$ images. To obtain high-quality scene graphs, we use all images in the training and testing sets to train the SGG model, and the validation set is used for selecting the best model. For scene graph embedding, we set a threashhold $0.2$ for relationship prediction, while we use the soft label of relation predictions as the relation-aware messages for reasoning. 

During HOI training, we set the initial learning rate to $0.01$ and dynamically decay it by 0.9 every 10 epochs.  To suppress the low confidence scores of detected objects, we use the same strategy proposed in previous work~\cite{li2019transferable}. As for the optimizer, we use Stochastic Gradient Descent (SGD) to optimize all parameters. All experiments are conducted on four Nvidia GeForce RTX 2080Ti GPUs with PyTorch.

\begin{table}[htb]
	\center
	
	\begin{tabular}{llr}
		\toprule
		\textbf{Methods} & \textbf{Feature Backbone} &  \textbf{mAP}$_\mathrm{role}$ \\ \midrule
		InteractNet~\cite{gkioxari2018detecting} & ResNet-50-FPN & 40.0 \\  
		GPNN~\cite{qi2018learning} &ResNet-152 & 44.0 \\  
		iCAN~\cite{gao2018ican} & ResNet-50 & 45.3 \\  
		TIK~\cite{li2019transferable} & ResNet-50 &47.8 \\  
		VSGNet~\cite{ulutan2020vsgnet} & ResNet-50 &51.1 \\  
		InterPoint~\cite{wang2020learning} &Hourglass-104 & 51.0 \\ 
		PDNet$^{\dagger}$~\cite{zhong2020polysemy} & ResNet-50 &51.6 \\  
		CHGNet~\cite{wang2020contextual} & ResNet-50 &52.7 \\ 
		FCNNet$^{\dagger}$~\cite{liu2020amplifying} & ResNet-50 &53.1 \\   \midrule
		\textbf{SG2HOI} &ResNet-50 & 52.8  \\  
		\textbf{SG2HOI}$^{\dagger}$ &ResNet-50 & \textbf{53.3} \\  
		\bottomrule
	\end{tabular}
	\caption{Performance comparison on the V-COCO dataset in terms of \textbf{mAP}$_\mathrm{role}$. The best score is marked in bold. $^{\dagger}$ denotes the models that use Faster-RCNN pre-trained on COCO~\cite{lin2014microsoft} as the feature extractor for human and object. }
	\label{tab.sota.vcoco}
	%
\end{table}

\begin{table*}[htb]
	\centering
	\begin{tabular}{ll*{6}{c}|c}
		\toprule
		\multicolumn{1}{l}{\multirow{2}{*}{Methods}} & \multicolumn{1}{c}{\multirow{2}{*}{Feature BackBone}} & \multicolumn{3}{c}{Default} & \multicolumn{3}{c }{Known Object} & \multicolumn{1}{l}{\multirow{2}{*}{Average}} \\  
		\multicolumn{1}{l}{} & \multicolumn{1}{l}{} & \multicolumn{1}{l }{Full} & \multicolumn{1}{l }{Rare} & \multicolumn{1}{l}{Non-Rare} & \multicolumn{1}{l }{Full} & \multicolumn{1}{l }{Rare} & \multicolumn{1}{l}{Non-Rare} & \\  \midrule
		InteractNet~\cite{gkioxari2018detecting} & ResNet-50-FPN & 9.94 & 7.16 & 10.77 & - & - & -& -\\
		GPNN~\cite{qi2018learning} & ResNet-152 & 13.11 & 9.34 & 14.23 & - & - & - &-\\
		iCAN~\cite{gao2018ican}  & ResNet-50 & 14.84 & 10.45 & 16.15 & 16.43 & 12.01 & 17.75 & 14.61\\
		TIK~\cite{li2019transferable}  & ResNet-50 & 17.03 & 13.42 & 18.11 & 19.17 & 15.51 & 20.26 & 17.25\\
		VSGNet~\cite{ulutan2020vsgnet} & ResNet-152 & 19.80 & 16.05 & 20.91 & - & - & -& \\
		InterPoint~\cite{wang2020learning} & Hourglass-104 & 19.56 & 12.79 & 21.58 & 22.05 & 15.77 & {23.92}  & 19.27 \\
		DPNet$^{\dagger}$~\cite{zhong2020polysemy} & ResNet-50 & 19.99 & 14.95 & 21.50 & 24.15 & 18.06 & \textbf{25.97} & 20.76\\
		CHGNet~\cite{wang2020contextual} & ResNet-50 & 17.57 & 16.85 & 17.78 & 21.00 & \textbf{20.74} & 21.08 & 19.18\\
		FCNNet$^{\dagger}$~\cite{liu2020amplifying} & ResNet-50 & 20.41 & 17.34 & 21.56 & 22.04 & 18.97 & 23.12 & 20.57\\ \midrule
		\textbf{SG2HOI} & ResNet-50 & 20.62 & 17.41 & 21.06 & 23.48 & 19.06 & 24.54 & 21.03 \\
		\textbf{SG2HOI}$^{\dagger}$ & ResNet-50 & \textbf{20.93}  & \textbf{18.24} & \textbf{21.78} & \textbf{24.83} & 20.52 & {{25.32}} & \textbf{21.94}\\
		
		\bottomrule
		
	\end{tabular}
	\caption{The comparison results on HICO-DET dataset in terms of Mean average precision (\textbf{mAP}). The best score is marked in bold. $^{\dagger}$ denotes the model uses Faster-RCNN pre-trained on COCO~\cite{lin2014microsoft} as the feature extractor for human and object. It is worth noting that InteractNet, GPNN, and VSGNet did not report their results on the Knowledge Object setting.   }
	\label{tab.sota.hico}
\end{table*}

\subsection{State-of-the-art Comparison}
We first present performance comparison  with nine recent state-of-the-art methods and report the mean Average Precision score results on both datasets.

Table~\ref{tab.sota.vcoco} shows  the comparison results    on the V-COCO dataset.  Among those methods, our SG2HOI gains competitive performance superior to the majority of methods. Specifically,  in the group  using the pretrained model on the ImageNet, SG2HOI surpasses the best model CHGNet~\cite{wang2020contextual} by $0.1$ percentage point. In fact, CHGNet is superior to other methods with the same feature extractor by at least $1.5$ points, possibly because  CHGNet also utilizes inter-class and intra-class feature refinements for human and object features.
However, as we discussed in Sec.~\ref{sec:ramp}, since CHGNet utilizes ambiguous messages to refine features and ignores the valuable relationship from the scene graph, 
its performance deteriorates a lot on the large and more complex HICO-DET dataset, which is shown in Table~\ref{tab.sota.hico}. 
On the other hand, when using the pre-trained Faster-RCNN with ResNet-50 backbone on COCO, we can observe a half point improvement (SG2HOI$^\dagger$ over SG2HOI). Besides, SG2HOI$^\dagger$  exceeds the SOTA model FCNNet by $0.2$ point. 

Table \ref{tab.sota.hico} shows the comparison results on HICO-DET. Following the experimental configurations in previous work~\cite{wang2020learning}, we also evaluate our model on three different HOI category sets: ``Full'', ``Rare'', and ``Non-Rare'', under two different schemes of ``Default'' and ``Known Objects''. As can be observed, our SG2HOI method consistently outperforms other state-of-the-art methods in both settings. Specifically, among the models that use a ResNet-50 model pre-trained on ImageNet as the feature extractor, our model achieves the best performance, with $0.76$ points higher than the best model InterPoint on average. 
Among the other group that use Faster-RCNN as the feature extractor, we could evidently observe SG2HOI$^\dagger$ is consistent superiority to the competitive models FCNNet and DPNet,  by $1.37$ and  $1.18$ percentages,respectively.

\subsection{Ablation Studies}
In this section we study the effectiveness of the two main components in our model: scene graph embedding and relation-aware message passing. To further compare to other similar strategies, we also test other two counterparts: image global feature from the last convolutional layer used in previous work~\cite{liu2020amplifying,ulutan2020vsgnet} and non-relation-aware message passing proposed in previous work~\cite{wang2020contextual}. 

Table \ref{tab.ab.1} shows the results on the V-COCO dataset, where the columns denote model variants with various modules included/excluded. The baseline column denotes the model that only uses visual appearance features and spatial information of humans and objects for HOI prediction. Rows sge, cov, rel, no-rel in Table~\ref{tab.ab.1} denote modules of scene graph embedding, {image global feature}, relation-aware message passing and non relation-aware message passing, respectively. Our full model is the last column (\ding{176}). 

\noindent \textbf{Scene Graph Embedding} aims at providing scene-specific global clues for HOI prediction, and the results in Table~\ref{tab.ab.1} show that this component (sge) can positively contribute to the performance improvement. Specifically, there is an increase of $2.8$ absolute points increase with the addition of scene graph embeddings (\ding{172} vs the baseline), and an increase of 5.7 points with the relation-aware message passing (\ding{173} vs the baseline). {Comparing {\ding{173}} to {\ding{174}}, we can see that the image global feature brings $0.9$ points lift. However, the performance of image global feature  is still $1.9$ points lower than our scene graph embedding. 
	Comparing {\ding{174}} with our full model {\ding{176}}, when switching from image features to scene graph embeddings, the model gains a $1.7$-point improvement. These results demonstrate that our proposed scene graph embedding module is not only effective but also superior to the conventional image features. 
	
	\begin{table}[]
		\centering
		\resizebox{\columnwidth}{!}{
			\begin{tabular}{l c c c c c c}
				\toprule
				Modules & baseline  &  \ding{172}   &  \ding{173}  &  \ding{174}  &   \ding{175}  &  \ding{176}   \\ \hline
				sge &  & $\checkmark$ &  &  & $\checkmark$ & $\checkmark$ \\  
				cov   &  &  &  & $\checkmark$ &  &  \\  
				rel  &  &  & $\checkmark$ & $\checkmark$ &  & $\checkmark$ \\  
				no-rel  &  &  &  &  & $\checkmark$ &  \\  \midrule
				\textbf{mAP}$_{\mathrm{role}}$& 46.5 & 49.3 & 50.2 & 51.1 & 51.0 & \textbf{52.8 }\\ \bottomrule
			\end{tabular}
		}
		\caption{The effectiveness of each component on the V-COCO dataset as measured by mean Average Precision (\textbf{mAP}$_\mathrm{role}$). Note that all the variant models use ResNet-50 pre-trained on ImageNet as the feature extractor. Compared with the baseline that only uses visual appearance and spatial features, our full model (\ding{176}) improves the performance by 13.55\%. }
		\label{tab.ab.1}
	\end{table}

	\begin{table}[htb]
		\centering
		\resizebox{\columnwidth}{!}{
			\begin{tabular}{l c c c c c c}
				\toprule
				Modules & baseline  &  \ding{172}   &  \ding{173}  &  \ding{174}  &   \ding{175}  &  \ding{176}   \\ \hline
				sge &  & $\checkmark$ &  &  & $\checkmark$ & $\checkmark$ \\  
				cov   &  &  &  & $\checkmark$ &  &  \\  
				rel  &  &  & $\checkmark$ & $\checkmark$ &  & $\checkmark$ \\  
				no-rel  &  &  &  &  & $\checkmark$ &  \\  \midrule
				\textbf{Default:}&   &   &   &   &   &  \\
				{Full}&   14.54 &  17.27 & 18.46  & 19.02  & 18.85  & \textbf{20.62 }\\
				{Rare}& 12.92 & 15.36 & 16.80 & 17.15 & {17.09} & \textbf{17.41 }\\
				{Non-Rare}& 16.37 & 17.48 & 19.14 & 19.50 & 19.02 & \textbf{21.06 }\\ \midrule
				\textbf{Known:}&   &   &   &   &   &  \\
				{Full}& 16.83 & 17.82 & 18.17 & 18.25 & 20.75 & \textbf{23.48 }\\
				{Rare}& 14.02 & 16.24 & 17.63 & 18.02 & 18.44 & \textbf{19.06  }\\
				{Non-Rare}& 18.01 & 19.25 & 19.70 & 20.07 & 22.31 & \textbf{24.54}
				\\
				\bottomrule
			\end{tabular}
		}
		\caption{The effectiveness of each component on the HICO-Det dataset, as measured by mean Average Precision (\textbf{mAP}$_\mathrm{role}$). Note that all the variant models use ResNet-50 pre-trained on ImageNet as the feature extractor. Compared with the baseline only using visual appearance and spatial features, our full model on average improves the performance by 33.1\%.}
		\label{tab.ab.2}
	\end{table}
	
	\begin{figure*}[htb]
		\centering
		\includegraphics[width=1\linewidth]{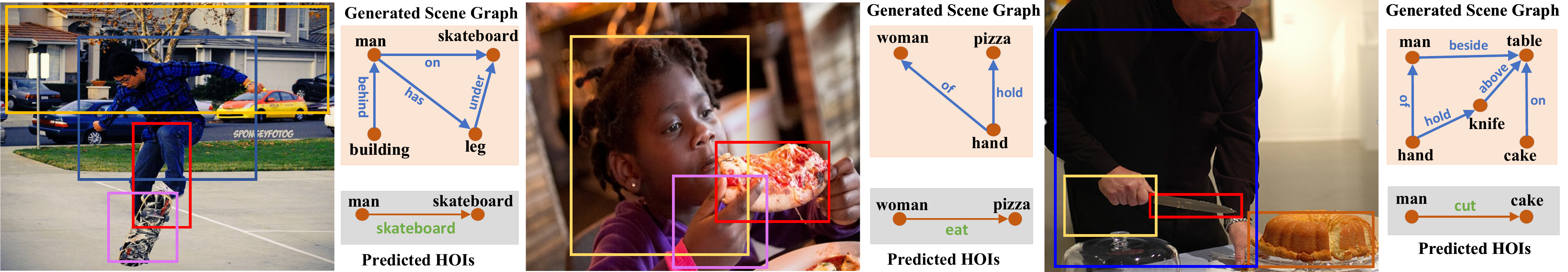}
		\caption{Human-Object Interaction detection and scene graph generation results on V-COCO. For each image, the top right graph is the generated scene graph which shows the basic relationships between object pairs, while the bottom right graph shows the predicted HOIs.  }
		\label{vis.vcoco}
	\end{figure*}

	\begin{figure*}[htb]
		\centering
		\includegraphics[width=1\linewidth]{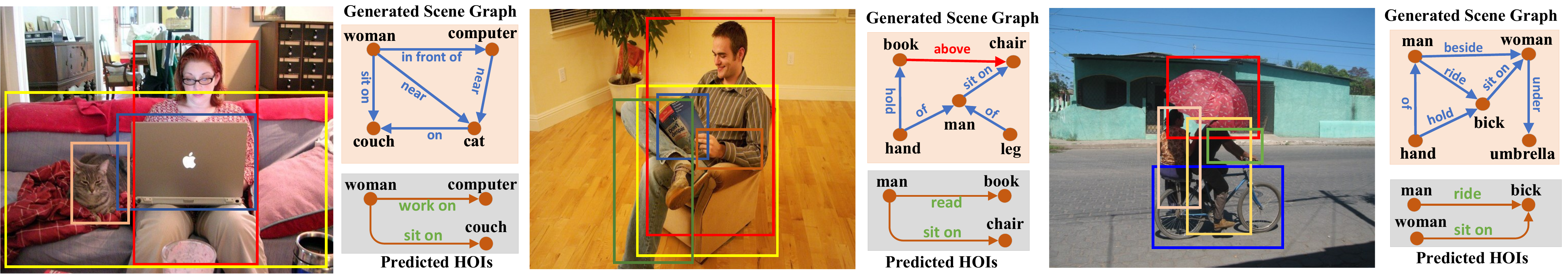}
		\caption{Human-Object Interaction detection and scene graph generation results on HICO-DET. Note that the red arrow for the middle image denotes the false predictions on both task.  }
		\label{vis.hico}
	\end{figure*}
	
	\noindent \textbf{Relation-aware Message Passing.} To evaluate this component, we design two baselines: (1) completely removing this module and (2) replacing it with the conventional non-relation-aware message passing strategy~\cite{wang2020contextual}. Comparing results of the baseline and \ding{173} in Table~\ref{tab.ab.1}, we can see there is a $3.7$ points improvement with the addition of the relation-aware message passing module. 
	Comparing \ding{172} with \ding{175}, we can observe that the addition of non-relation-aware message passing contributes $1.7$ points increase. However, non-relation-aware message passing still lags behind our full model by $1.8$ percentage points (\ding{175} vs \ding{176}). 
	
	We also conduct the ablation experiments on the HICO-DET dataset, and the results are presented in Table \ref{tab.ab.2}, from which we could obtain the same conclusions as on the V-COCO dataset.

	\subsection{Qualitative Results}
	
	We further visualize some examples on both datasets, as shown in Figures~\ref{vis.vcoco} and~\ref{vis.hico}, respectively. For each image, we draw its scene graph and predicted HOIs.  
	
	Generally, we could find that the generated scene graph is more detailed than the HOI graph, and the two graphs have intrinsic correlations. Taking the first image in Figure~\ref{vis.vcoco} as an example,  as long as we know a man is on a skateboard and his leg is \texttt{under} the skateboard, we are likely to guess the interaction is skateboard, because the majority of skateboarding people have the same relationships.  
	
	For the example from the HICO-DET shown in Figure~\ref{vis.hico}, some SG relation are the same as interaction classes, such as \texttt{sit on} and \texttt{ride}. Therefore, we could directly obtain these interactions. Additionally, there is a wrong relationship predicted in the scene graph: in fact, there is no relation between book and chair in the middle image in Figure~\ref{vis.hico}. Interestingly, since the SGG model predicts the relation between hand and book as \texttt{hold}, the HOI model also falsely predicts the interaction as \texttt{hold}. Thus, the quality of scene graph generation has an effect on the prediction of HOIs.

\section{Conclusion}

In this paper, we propose a novel framework, denoted as SG2HOI, that utilizes scene graph information as the key contextual cues to predict Human-Object Interaction. To the best of our knowledge, our method is the first to bridge the gap between the two tasks. Specifically, we achieve this goal from two aspects. First, we embed the global scene graph of each image as the scene-specific context, for which we propose two embedding strategies: scene graph layout embedding and attention-based relation fusion. Secondly, we treat each scene graph as a reasoning graph and devise a novel relation-aware message passing mechanism that gathers the relationship information from its inter-class and intra-class neighbors. 
We conduct a wide range of evaluation on two benchmark datasets: V-COCO and HICO-DET, which shows that our model's performance is superior to current state-of-the-art methods. In the future, we will integrate the two tasks as a dual-task learning problem to improve the performance of both tasks. 
	{\small
		\bibliographystyle{ieee_fullname}
		\bibliography{egbib}
	}
	
\end{document}